# Roulette-Wheel Selection-Based PSO Algorithm for Solving the Vehicle Routing Problem with Time Windows

Gautam Siddharth Kashyap[1], Alexander E. I. Brownlee[2], Orchid Chetia Phukan[3], Karan Malik[4], Samar Wazir[5]

[1]officialgautamgsk.gsk@gmail.com, [2]sbr@cs.stir.ac.uk, [3]orchidp@iiitd.ac.in, [4]karanmalik2000@gmail.com, [5]samar.wazir786@gmail.com,

[1,5]Department of Computer Science and Engineering, SEST, Jamia Hamdard, New Delhi, India
[2]Division of Computing Science & Mathematics, University of Stirling, UK
[3]IIIT-Delhi, New Delhi, India
[4]Arizona State University, Tempe, Arizona, USA

*Abstract*— The well-known Vehicle Routing Problem with Time Windows (VRPTW) aims to reduce the cost of moving goods between several destinations while accommodating constraints like set time windows for certain locations and vehicle capacity. Applications of the VRPTW problem in the real world include Supply Chain Management (SCM) and logistic dispatching, both of which are crucial to the economy and are expanding quickly as work habits change.

Therefore, to solve the VRPTW problem, metaheuristic algorithms i.e. Particle Swarm Optimization (PSO) have been found to work effectively, however, they can experience premature convergence. To lower the risk of PSO's premature convergence, the authors have solved VRPTW in this paper utilising a novel form of the PSO methodology that uses the Roulette Wheel Method (RWPSO). Computing experiments using the Solomon VRPTW benchmark datasets on the RWPSO demonstrate that RWPSO is competitive with other state-of-the-art algorithms from the literature. Also, comparisons with two cutting-edge algorithms from the literature show how competitive the suggested algorithm is.

## I. INTRODUCTION

The Vehicle Routing Problem (VRP) is a well-known problem that seeks to reduce the cost of moving goods by creating an effective strategy for vehicle routing between depots and customers. The VRPTW is an extension of VRP that establishes time slots for visiting particular locations. VRPTW has a variety of practical applications, including bank and mail deliveries, security patrol services, school bus routing, etc. Many VRP problems go beyond measuring only the total journey cost and separately seek to optimize the number of vehicles and the distance travelled. According to the VRP problem structure and the constraints of VRPTW, the improvement of one objective may conflict with the other objectives. Therefore, this makes the VRPTW a multi-objective search problem [1].

It has been noted that PSO has found some success in solving the VRPTW problems, such as [2]–[8], as it has several advantages, including the fact that the basic algorithm is very easy to implement and provides quick and reliable results. However, premature convergence can happen in PSO for complex multi-objective search problems, just like in other stochastic optimization algorithms like Genetic Algorithms (GA) [9]. The fact that the best particle discovered by a swarm frequently falls into local minima is one of the main causes of premature convergence in PSO. To enhance the performance of PSO, it has been proposed to use chaotic maps, incorporate mathematical or stochastic operators, local searches, or hybridise the algorithms. The hybridise algorithm has movements that are notably more diversified than the traditional PSO algorithms. As a result, it avoids local optima and looks into different areas of the search space for a problem more extensively. Therefore, in this paper, the Roulette Wheel Method has been deployed based on their fitness value to choose the search agents for the PSO and take advantage of promising parts of the search space. As a result, the main contribution of this paper is the suggestion of a hybrid PSO algorithm using the Roulette Wheel Method, which is referred to as Roulette Wheel Particle Swarm Optimization (RWPSO). In RWPSO, the best particle is not necessarily the one with the highest fitness value. Instead, the authors use the Roulette Wheel Method to stochastically choose the particle from the population. According to the hypothesis that occasionally choosing fewer suited individuals might sometimes increase efficiency, therefore, the RWPSO is less likely to fall into local minima by weakening the strong bias towards the global optimum [10].

The paper is organized as follows. In the next section, a literature review of VRPTW is presented. In Section III, the background and mathematical model for VRPTW are discussed. In Section IV, the background of PSO followed by its extension to RWPSO is discussed. In Section V, the authors show how RWPSO is applied to VRPTW. In Section VI, computational experiments on benchmark problems demonstrate the utility of the algorithm and the authors conclude the paper in Section VII.

## II. LITERATURE REVIEW

The VRP, developed by Dantzig & Ramser [11], is a crucial idea in the SCM system, in which a specific product must be delivered to a specific customer via a specified route. They discovered a path to deliver fuel to gas stations in their initial mathematical programming model. A heuristic algorithm for VRP was later proposed by Clarke & Wright [12]. Based on the types and constraints of VRP, a lot of research has been conducted in this area. The authors now examine recent research that is pertinent to VRPTW, organised according to the methodology.

### A. Exact Method

Typically, the Mixed Integer Linear Programming (MILP) model is used to solve the VRPTW problems exactly. The VRPTW with pick-up and delivery was resolved by Dumas et al. [13] by suggesting an exact algorithm with column generation. Jeroen [14] uses the mathematical model to produce an exact solution for the VRPTW. To solve the VRPTW, Taş et al. [15] devised a linear programming methodology. By combining two exact solution algorithms, Hernandez et al. [16] developed an algorithm for VRPTW with multiple trips of the vehicle. The branch and price concept, which consists of two different kinds of sets that cover the formulations, was employed by the researchers. They demonstrated the effectiveness of their algorithm using a few benchmark problems. Alvarez & Munari [17] solved the VRPTW with service time at customers based on several delivery men allotted to the routes to serve them. The researchers combined the branch-and-price concept as well as the cut algorithm and two metaheuristic algorithms. Parragh & Cordeau [18] proposed an algorithm for VRPTW with trucks and trailers for vehicle groups by using a branch-and-price concept. The adaptive large neighbourhood search algorithm was also used to compute initial columns. The algorithm was examined on some benchmark problems and showed competitive results.

### B. Heuristic Method

Heuristic algorithms have also been employed to solve the VRPTW problem such as Schulze & Fahle [19] used a parallel Tabu Search (TS) algorithm. The VRPTW was resolved by Barán & Schaerer [20] using Ant Colony Optimization (ACO) algorithms. Le Bouthillier & Crainic [21] solved the VRPTW by utilizing a cooperative parallel metaheuristic, combining exact and heuristic algorithms. Ho & Haugland [22] solved the VRPTW with split deliveries using the TS algorithm. Using a GA with crossover operators, Nazif & Lee [23] solved the VRPTW. Additionally, they examined their suggested algorithm on a few benchmark problems and compared it to other algorithms that had similar performance. The VRPTW with a heterogeneous fleet was solved by Jiang et al. [24] using the TS algorithm. By utilising a GA with a new crossover operator, Pierre & Zakaria [25] were able to solve the VRPTW. In addition to producing competitive results on benchmark problems when compared to a traditional GA, they found that this new crossover operator enhances the local performance of the GA. To address the multi-objective VRPTW, Chiang & Hsu [26] suggested a heuristic algorithm based on a universal algorithm, to reduce both the number of vehicles and the overall distance travelled. The ship routing problem and the scheduling problem with time window limitations were both solved by De et al. [2] using PSO with composite particles. The problem was presented as a Mixed Integer Non-Linear Programming problem (MINLP). Iqbal et al. [27] solved a multi-objective VRPTW by using a hybrid metaheuristic based on an Artificial Bee Colony (ABC) algorithm. Using a heuristic algorithm, Cetin & Gencer [28] addressed the VRP based on simultaneous pickup and delivery as well as time windows. Four different forms of VRPTW with a heterogeneous fleet were solved by Koç et al. [29] using a hybrid evolutionary algorithm. The experiments produced positive results as they attempted to reduce the route time and distance travelled. Tan et al. [30] solved the VRPTW by using Bacterial Foraging Optimization (BFO) algorithm. They claimed that their suggested algorithms outperformed the GA and PSO in terms of performance. Yan et al. [31] solved the split delivery VRPTW by proposing a two-step solution algorithm. Bae & Moon [32] used a heuristic algorithm using a MILP model to solve the VRPTW with numerous depots while attempting to reduce the costs of the delivery route and vehicle installation. According to the researchers, their suggested algorithm successfully solves complicated problems. To solve the VRPTW while taking into constraints like carbon emission, fuel usage, and fuel costs, De et al. [3] employed the PSO-CP algorithm with an MINLP model. They demonstrated competitive results by comparing their suggested algorithm with PSO and a GA. Using a multi-depot and heterogeneous fleet, Dondo & Cerdá [33] solved the VRPTW using a cluster-based optimization algorithm. To solve the VRPTW while taking into constraints like synchronisation and precedence, Haddadene et al. [34] employed a greedy heuristic algorithm with a MILP model. The researchers demonstrated that their proposed algorithm is faster on small and medium datasets. The maritime inventory routing problem with time windows was resolved by De et al. [4] using a PSO-CP algorithm. Miranda & Conceiço [35] proposed a new metaheuristic algorithm that makes use of the iterated local search algorithm to solve the VRPTW with stochastic travel and service times. In Solomon datasets [36], they examined their algorithm; it performed better than competing heuristic algorithms.

Toth & Vigo [37] provide a solid, albeit somewhat outdated, overview of VRP. Since it gets more complex to handle VRP problems with exact algorithms as they grow in size, the research mentioned above has tended to favour heuristic algorithms. Occasionally in the past, like in [38], [39], the idea of including Roulette Wheel Method into PSO has been floated, albeit there aren't many instances of it being used to solve VRP problems. The only two appear to be [40]- but in this case, the approach was applied to VRP aimed at minimising cost only, and with no time windows and [41]- applied Roulette Wheel Method in the discretisation step, rather than the selection of *gbest* as the RWPSO do here. The authors made an effort to highlight the research that was pertinent to the strategy of RWPSO. According to our survey, no PSO methodology has ever attempted a stochastic method of selecting the *gbest*, as the RWPSO do here. To prevent premature convergence, the goal of this paper is to select the *gbest* by using the RWPSO algorithm. As a result, the authors think that our research on VRP will be viewed as a unique contribution.

### III. Background of VRPTW

The VRP was first proposed by Dantzig & Ramser [11]. The objective is to determine the best routes between the depots and the customers in a specific geographic area so that items can be delivered effectively while meeting a set of constraints. VRP has a substantial impact on logistics and distribution, with common concrete applications including the delivery of mail and food items. A variant of the VRP where customers must be supplied within a specific window of time is VRPTW. Finding a solution that: 1) Reduces the number of vehicles needed, 2) Reduces the average travel time, and 3) Reduces the average distance travelled by vehicles is the major goal of VRPTW. Two varieties of VRPTW are:

1. **Vehicle Routing Problem with Hard Time-Windows (VRPHTW):** A vehicle must wait at the customer's location if it arrives before the customer's scheduled service time; nevertheless, there is no additional cost. The customer cannot be served if the vehicle arrives after the service time.
2. **Vehicle Routing Problem with Soft Time-Windows (VRPSTW):** An additional cost (i.e., a penalty) will be charged to the service if a customer is visited outside the allotted time [42].

Since VRPTW is NP-Hard, it is complex to find exact solutions for larger instances. For the several versions of VRP, numerous heuristic and metaheuristic algorithms have been developed (as discussed in Section II). PSO, a metaheuristic algorithm inspired by flocks of birds is used in this paper and has shown to be effective for a variety of multi-objective problems, but less effective for discrete and combinatorial problems (as discussed in Section I). The mathematical models for VRPTW will now be discussed in the following sub-section, and then we'll move on to the RWPSO algorithm.

#### A. Mathematical Model for the VRPTW

A group of homogeneous vehicles denoted as *vec* having restricted capacity $LC$ serves the VRPTW. From a graph-theoretical point of view, Chen et al. [43] described VRPTW as follows:-

Let an undirected graph *G* be:

$$G = (V, E) \tag{1}$$

where $V$ is a set of vertices and $E$ is a set of edges. Let each vertex $v_i$ $(i = 0, 1, 2 ..., n)$ represent a served customer $v_0$ related to the depot. The connection between the depots and customers and vice-versa is represented by the set of edges. And, let the demands of customers $i$ be $c_i$, the service time be $s_i$ and time windows be:

$$[start_i, end_i] \tag{2}$$

A vehicle must reach customer $i$ before $end_i$. The vehicle can reach $v_i$ before $start_i$; the implications depend on whether it is VRPSTW or VRPHTW. The depot maintains a time-window $[start_0, end_0]$: this time interval is called the *scheduling horizon* which means that the vehicle cannot leave the depot before $start_0$ and has to return to the depot on or before $end_0$. Let there be another special time window $[start\_traffic, end\_traffic]$ to avoid servicing the customer during heavy traffic hours. Here $start\_traffic$ and $end\_traffic$ mean the starting and ending times of traffic in the traffic route. Traffic police and control towers can provide this time. It is assumed that the symmetric distance matrix $DM = (d_{ij})$ satisfies the triangle inequality. Travel time $tt_{ij}$ is directly proportional to $d_{ij}$. Two decision variables, *x* and *y*, are also defined. For each edge $(i,j)$, where $i \neq j$ and $i,j \neq 0$ and each vehicle *vec,* we define $x_{ijvec}$ by:

$$x_{ijvec} = \begin{cases} 0; & if\ vec\ does\ not\ drive\ from\ vertex\ i\ to\ vertex\ j \\ 1; & otherwise \end{cases} \tag{3}$$

A decision variable $y_{ivec}$ is defined for each vehicle *vec* at each vertex *i* for serving the customer *i* by time vehicle *vec*. In case the vehicle *vec* does not service the customer *i* then $y_{ivec}$ is called off. The authors assume that $start_0 = 0$ and therefore $y_{0vec} = 0$, for all *vec*. The objective of the problem is to find an optimal transportation cost for one vehicle such that: 1) Each customer should be served only once, 2) Each route will start & end from vertex 0, and 3) The time windows and capacity constraints should also be respected. Mathematically the VRPTW used by [36] is as follows:

$$\min z = \sum_{vec \in V} \sum_{i \in N} \sum_{j \in N} c_{ij} \times x_{ijvec} \tag{4}$$

Subject to:-

$$\sum_{vec \in V} \sum_{j \in N} x_{ijvec} = 1 \qquad \forall i \in N \tag{5}$$

$$\sum_{i \in N} cust_i \sum_{j \in N} x_{ijvec} \leq LC \qquad \forall vec \in V \tag{6}$$

$$\sum_{j \in N} x_{0jvec} = 1 \qquad \forall vec \in V \tag{7}$$

$$\sum_{i \in N} x_{ihvec} - \sum_{j \in N} x_{hjvec} = 0 \qquad \forall h \in N, \forall vec \in V \tag{8}$$

$$\sum_{i \in N} x_{i0vec} = 1 \qquad \forall vec \in V \tag{9}$$

$$y_{ivec} + s_i + tt_{ij} - K(1 - x_{ijvec}) \leq s_{jvec} \qquad \forall i \in N \tag{10}$$

$$p_j > \varepsilon * (start\_traffic - tt_{ij}) * (end\_traffic - tt_{ij}) \qquad \forall j \in j \tag{11}$$

$$start_i \leq y_{ivec} \leq end_i \qquad \forall i \in N, \ \forall vec \in V \tag{12}$$

$$x_{ijvec} \in \{0,1\} \qquad \forall i,j \in N, \forall vec \in V \tag{13}$$

where Constraint (5) means that the customer is visited exactly once; Constraint (6) means that the vehicle is loaded within the capacity; Constraints (7) to (9) mean that the vehicle will leave the depot for the customer and will also ensure that after serving the customer the vehicle will return to the depot; Constraint (10) implies that if a vehicle *vec* travels from customer *i* to customer *j* but it cannot arrive at customer *j* before $y_{ivec} + s_i + tt_{ij}$, where $K$ is a large scalar; Constraint (11) defines the penalty to the objective function if the vehicle is located at the traffic time window; Constraint (12) ensures that the time windows are respected and Constraint (13) is an integrality constraint.

## IV. PSO AND EXTENSION TO RWPSO

Kennedy & Eberhart [44] developed the well-known optimization population-based algorithm named as PSO. The behaviour of a swarm of bees, a flock of birds, and other groupings of autonomous entities with a social component and a common purpose served as the basis for the PSO. As a search tool, it keeps track of a swarm's individuals' physical motion as well as their mental and social activities. In PSO, the problem solution is represented by the position of a multidimensional particle, and the particle velocity is a representation of the particle searching capability. In general, Algorithm 1 depicts the fundamental version of PSO. When a particle achieves a position with a greater fitness value than the fitness value of the previous local best, the cognitive behaviour, which represents the particle's personal best position, is updated. The population's best position is determined by the social behaviour of each individual particle, which indicates the particle's global best position. This social behaviour is changed anytime a particle achieves a position with a better fitness value than the previous global best. The PSO algorithm can be defined more explicitly as follows:

In $N$-dimensional space let there be an $m'$ number of particles. Therefore:- Let $\mathbb{X}_i = [\mathbb{x}_{i1}, \mathbb{x}_{i2}, \mathbb{x}_{i3}, \dots \mathbb{x}_{iN}]^T$ represent the position of the $i^{th}$ particle; Let the highest fitness value given by the best position be: $\mathbb{P}_i = [\mathbb{p}_{i1}, \mathbb{p}_{i2}, \mathbb{p}_{i3}, \dots \mathbb{p}_{iN}]^T$; Let the best position found by the population be: $\mathbb{P}_g = [\mathbb{p}_{g1}, \mathbb{p}_{g2}, \mathbb{p}_{g3}, \dots \mathbb{p}_{gN}]^T$; The $i^{th}$ particle moves to the search space with a velocity: $\mathbb{V}_i = [\mathbb{v}_{i1}, \mathbb{v}_{i2}, \mathbb{v}_{i3}, \dots \mathbb{v}_{iN}]^T$. Then the updated velocity and position of the $i^{th}$ particle will be:-

$$\mathbb{v}_{in}(t+1) = w * \mathbb{v}_{in}(t) + c_1 * r_1(\mathbb{p}_{in}(t) - \mathbb{x}_{in}(t)) + c_2 * r_2(\mathbb{p}_{gn}(t) - \mathbb{x}_{in}(t)) \quad (14)$$

$$\mathbb{x}_{in}(t+1) = \mathbb{x}_{in}(t) + \mathbb{v}_{in}(t+1) \quad (15)$$

where $c_1 \& c_2$ are two positive constants; $r_1 \& r_2$ are two random values in the range *[0,1]* and $w$ is the inertia weight. To prevent a particle from leaving the search space, the velocity vector $V$ is clamped to the range $[-V_{max}, V_{max}]$ and the position vector $X$ is clamped to the range $[X_{min}, X_{max}]$ during each iteration. $X_{min} \& X_{max}$ are chosen according to the problem and $V_{max}$ is usually chosen to be $k \times X_{max}$, with $0.1 \leq k \leq 1.0$ [45]. However, in PSO, premature convergence can occur. Therefore, premature convergence can be detected by monitoring the extrema $\mathbb{P}_i \& \mathbb{P}_g$; if they become equal then all particles will quickly approach to $\mathbb{P}_g$ and premature convergence occurs.

1. *Initialise the population (swarm) with each particle having a random position and velocity.*
2. *Each iteration until stopping criterion is met:*
   2.1. *Modify each particle's velocity based on the particle's own best known position and swarm's global best position*
   2.2. *Move each particle to a new position according to its velocity*
   2.3. *Evaluate quality of each particle*
   2.4. *Update the swarm's and each particle's best known positions*

**Algorithm 1. The overall framework of PSO.**

### A. Roulette Wheel Particle Swarm Optimization (RWPSO)

To avoid premature convergence, $\mathbb{P}_g$ should be prevented from falling into local minima, or at least have the probability of this reduced. In this section, the novel approach the authors adopt is that the selection scheme is improved by employing the Roulette Wheel Method, well-established in GA, in which the global "*best*" particle is chosen probabilistically from a group of particles. Therefore:- Let the position of the first $L$ particles with the highest fitness be $\mathbb{P}_g{}^i = (\mathbb{P}_g{}^1, \mathbb{P}_g{}^2, \mathbb{P}_g{}^3, \dots \dots, \mathbb{P}_g{}^L)$ and their fitness by $g^i = (g^1, g^2, g^3, \dots \dots, g^L)$ where $(i = 1,2,3, \dots, L)$.

According to the Roulette Wheel Method, the probability of $\mathbb{P}_g{}^i (i = 1,2,3, \dots, L)$ being selected as $\mathbb{P}_g$ is:-

$$O^i = g^i / \sum_{j=1}^{L} g^j \ (i = 1,2,3, \dots, L) \quad (16)$$

1. *According to the problem: set the number of iterations, determine the position vector $[X_{min}, X_{max}]$ and velocity vector $[-V_{max}, V_{max}]$ respectively. Initialize the position and velocity of $m'$ particles randomly and calculate their fitness $\mathbb{F}_i$.*
2. *Let the best position found by the $i^{th}$ particle be $\mathbb{P}_i = \mathbb{X}_i$ and its fitness be $\mathbb{F}best_i = \mathbb{F}_i$ and let the position of the first $L$ particles with the highest fitness be $\mathbb{P}_g{}^i = (\mathbb{P}_g{}^1, \mathbb{P}_g{}^2, \mathbb{P}_g{}^3, \dots \dots, \mathbb{P}_g{}^L)$ and their fitness be $g^i = (g^1, g^2, g^3, \dots \dots, g^L)$ where $(i = 1,2,3, \dots, L)$ and satisfy $g^1 \geq g^2 \geq g^3 \geq \cdots \geq g^L$.*
3. *Calculate the probability of $\mathbb{P}_g{}^i = (\mathbb{P}_g{}^1, \mathbb{P}_g{}^2, \mathbb{P}_g{}^3, \dots \dots, \mathbb{P}_g{}^L)$ as $\mathbb{P}_g$ according to (20).*
4. *Generate a random value $\xi$ in the range [0,1] therefore: If $\xi$ satisfies $0 \leq \xi \leq O^1$ then let $\mathbb{P}_g = \mathbb{P}_g{}^1$. Else if $\xi$ satisfies $\sum_{j=1}^{k-1} O^j \leq \xi \leq \sum_{j=1}^{k} O^j$ then let $\mathbb{P}_g = \mathbb{P}_g{}^k (k = 2,3,4, \dots, L)$.*
5. *Update the position and velocity using equations (17 & 18) and restrict the velocity and position vectors to the range $[-V_{max}, V_{max}]$ and $[X_{min}, X_{max}]$ respectively.*
6. *if $\mathbb{F}_i > \mathbb{F}best_i$ then let $\mathbb{P}_i = \mathbb{X}_i$ and $\mathbb{F}best_i = \mathbb{F}_i$.*
7. *if $\mathbb{F}_i > g_1$ then let $\mathbb{P}_g{}^1 = X_i$ and $g^1 = \mathbb{F}_i$. Else if $g^{j-1} > \mathbb{F}_i > g^j (j = 2,3,4, \dots, L)$ then let $\mathbb{P}_g{}^j = X_i$ and $g^j = \mathbb{F}_i$.*
8. *If the maximum number of iterations is reached then stop, otherwise go back to step 3.*

**Algorithm 2. RWPSO.**

As a result, RWPSO maintains the fundamental properties of PSO, i.e., RWPSO searches for an ideal solution while also keeping an eye on the two extrema. The likelihood of premature convergence is reduced when the best particle falls into a local minimum because a new particle may be chosen as $\mathbb{P}_g$ and the particles move in opposite directions. The particle will go in a generally positive path even when the found one is not the best. The RWPSO implementation is displayed in Algorithm 2.

## V. APPLICATION OF RWPSO TO VRPTW

The initialization of the particles starts the RWPSO algorithm. A different particle may be chosen as $\mathbb{P}_g$ and moved to the optimal place through a sequence of steps if the best particle falls into a local minimum. It is noteworthy that the emphasis is on selecting the ideal position for the particle. So, to implement the RWPSO-VRPTW, the relationship between the particle's position and the problem solution must be established. The particle position is converted to the solution to the problem via a decoding procedure.

### A. Pseudocode of RWPSO-VRPTW

In Algorithm 2, the authors presented the RWPSO. The authors now go over the implementation, and Algorithm 3 provides a more condensed form of the suggested algorithm to help with the discussion. In the subsections that follow, the authors discuss each stage.

1. *Initialize the particles.*
2. *Decode the particles.*
3. *Find the route performance of each particle.*
4. *Update the parameters according to Algorithm 2.*
5. *Move the particle to the next position.*
6. *Check the stopping condition:*
7. *If the stopping condition is met, then end the program.*
8. *If the stopping condition is not met, then go back to step 2.*

**Algorithm 3. RWPSO-VRPTW.**

*1) Initialization:* The velocity, position, and personal best values for the $i^{th}$ particle are set in this phase to initialize the particle. The position value for each dimension of a particle is set at random according to the range of the position vector $[X_{min}, X_{max}]$. The personal best value is equal to the position value and the velocity is set to 0.

*2) Decoding of the Particle & Route Performance:* The decoding method is the second step of the RWPSO-VRPTW framework. The main task of this method is to turn each particle into a series of vehicle routes, with the output of the particle serving as the fitness value. An indirect representation of the VRPTW solution is present in Fig. 1. In *(c+2\*vec)* dimensional, let the number of the customers be *cust* and let the number of vehicles be *vec*.

| 1 | 2 | ... | c | c+1 | c+2 | c+3 | c+4 | ... | c+2*vec-1 | c+2*vec |
|---|---|---|---|---|---|---|---|---|---|---|
| Customer Priority | | | | Vehicle 1 | | Vehicle 2 | | | Vehicle vec | |

Fig. 1. Solution encoding

In each dimension, a real number represents the particle position. The authors have chosen an indirect representation since continuous integers cannot be used to indicate vehicle routes directly. The following are the two primary parts of this representation:

1. In the first section of the representation, the customers are represented by increasing numbers, with the first customer having a higher priority than the second. To represent the *n* number of customers, *n* dimensions are needed. As a result, it is possible to create an unlimited number of combinations of *n* continuous numbers, each of which results in a single priority list that generates a variety of vehicle routes.
2. The vehicle reference point is discussed in the second section of the representation. The position of the customer that the vehicle must serve is referred to as the reference point.

The decoding procedure works as follows:

1. At the initial step of the decoding procedure, the priority list is established based on the particle position. The small position value principle, which states that the higher the value assigned to the customer, the lower the position value; is the foundation upon which the priority list is constructed.
2. The corresponding dimension is changed into a reference point in the second phase.

*3) Update the Parameters & Move the Particle to the Next Position:* Section III of the RWPSO-VRPTW framework has explained this stage.

*4) Stopping Condition:* The stopping condition controls how many times the algorithm iterates, preventing the method from going indefinitely roundabout. The RWPSO-VRPTW framework uses the total number of iterations as the stopping condition.

## VI. COMPUTATIONAL EXPERIMENT

Computational experiments are conducted to evaluate the effectiveness of the RWPSO using [36] benchmark datasets. Our algorithm is put into practice in Google Colab using an AMD E1-6010 APU running at 1.35 GHz with 4GB RAM. Table I contains a list of the RWPSO algorithm's parameters.

Table I. Parameters of the RWPSO

| Parameters | Value |
|---|---|
| Number of Particles | 20 |
| $c_1$ | 2.0 |
| $c_2$ | 2.0 |
| Inertia Factor | 0.9 (linearly decreasing to 0.4 during the training process) |
| Rate | 2.0 |
| Time Cost | 10 |
| Penalty Cost | 100 |

*Note: The parameters listed above were obtained from studies [5] and [6] and are equivalent for comparison purposes.*

## A. Benchmark Data

The well-known [36] benchmark datasets have been employed in our experiments. Heap Distribution (C), Random Distribution (R), and Semi-Heap Distribution (RC) are the three types of customer datasets that [36] created. Customers have grouped according to heap allocation. The established vehicle routes are used to construct time windows from these clustered customers. In a random distribution, the randomly produced customers are distributed evenly across a square, however, in a semi-heap distribution, the customers are mixed. The datasets C1, R1, and RC1 also have minimal delivery capacities and constrained time frames, which limit the number of customers that can be handled by a single vehicle. The datasets C2, R2, and RC2 on the other hand, offer enormous delivery capacities and time periods that enable numerous customers to be served by a single vehicle. The generation of these datasets has an impact on how the routing and scheduling algorithm behaves. The following variables can have an impact on the algorithm: 1) Geographical information, 2) The number of customers the vehicle serves, 3) The percentage of customers with time limits, and 4) The severity and location of the time frames. There are a total of 56 instances, including 17 C instances, 23 R instances, and 16 RC instances. In this paper, benchmarks for 25 customers, 50 customers, and 100 customers are examined. According to the benchmarks provided by the [36], the iteration count has been set at 1000 for the first 25 customers, and at 10,000 for the next 50 and 100 customers. The RWPSO algorithm's results are compared with the most recent top results, which are reported in [46]. The authors compared the RWPSO with other cutting-edge algorithms, such as [5] and [6], in addition to comparing our results to those described in [46]. Similar to [5] and [6], the main goal of optimization in these experiments is to minimise the number of vehicles, with the reduction in distance travelled serving as a secondary goal. As a result, based on the number of vehicles, [5] and [6] are compared to the RWPSO.

## B. Results and Comparison

The comparison of the results is shown in Tables II to IV, where NV denotes the number of vehicles, TD denotes the total distance travelled, and average CPU time denotes the computational time. The tables also include the best, mean, and standard deviation results. Using the [5] convention, (*) means that RWPSO achieved the best or nearly the best result for TD on the same number of vehicles, while (**) indicates that RWPSO achieved the best result while using fewer vehicles. Within 5% of the best-known result is what is referred to as the near-best result for TD. RWPSO is compared to two studies ([5] and [6]) and to PSO in Tables II, III, and IV.

*1) 25 Customer Instances:* As shown in Table II, for the 25 customer instances, the RWPSO achieved 28 best or near-best TD results on the same number of vehicles (*); and 28 results matching or improving on the number of vehicles (**). Table II demonstrates that 28 (*) C101, C102, C103, C104, C105, C106, C107, C108, C109, R201, R101, R102, R104, R106, R107, R108, R110, R112, R208, RC101, RC102, RC103, RC104, RC105, RC016, RC107, and RC108 while 28 (**) emerged for the C202, C203, C204, C205, C206, C207, C208, R103, R105, R109, R111, R201, R202, R203, R204, R205, R206, R207, R209, R210, R211, RC201, RC202, RC203, RC204, RC205, RC206, and RC207. The degree to which the actual results match the predicted results can be determined by a small mean difference. A limited range between the lowest and greatest values of the target function, once more, demonstrates stability between repetitions. The fact that this result ensures that a workable solution will be found in a reasonable length of time makes it a desirable one as well. The results indicate that, in the 25 customer instances, which were the consequence of various problem scenarios, the RWPSO frequently offers a workable solution.

*2) 50 Customer Instances:* As shown in Table III, for 50 customer instances, the RWPSO finds 29 best or near-best results on the same number of vehicles (*); and 28 match the number of vehicles (**). In Table 3, 27 (*) C101, C102, C103, C104, C105, C106, C107, C108, C109, C204, C208, R104, R106, R108, R110, R112, R204, R207, R208, RC101, RC102, RC103, RC104, RC105, RC016, RC107, RC108, RC207and RC208, while 28 (**) appears for the following values: C201, C202, C203, C205, C206, C207; R101, R102, R103, R105, R107, R109, R111; R201, R202, R203, R205, R206; R209, R210, R211; RC201, RC202, RC203, RC204, RC205, and RC206. There isn't a certain algorithm that consistently yields the best results. Bold results in Table III denote the new best values that the RWPSO has generated. The PSO response level is effective for the 50 customers, as indicated in Table III. The average variance is a certain percentage, assuming that the actual results are relatively close to the optimal response. The results demonstrate that the applied RWPSO typically produces an adequate result because the 50 customer instances were the outcomes of varied circumstances.

*3) 100 Customer Instances:* As shown in Table IV, for 100 customer instances, the RWPSO has 41 best or near-best results on the same number of vehicles (*); with 15 results matching the number of vehicles (**). Table IV shows that while 41 (*) matched or reduced the value of TD i.e. C102, C103, C104, C106, C109, C201, C202, C206, C207, C208, R103, R104, R105, R106, R107, R108, R109, R110, R111, R112, R201, R202, R203, R205, R206, R207, R208, R209, R210, R211, RC101, RC105, RC106, RC107, RC202, RC203, RC204, RC205, RC206, RC207, and RC208 while 15 (**) utilised in C101, C105, C107, C108, C203, C204, C205, R101, R102, R204, RC102, RC103, RC104, RC108, and RC201 still surpassed the minimum value of TD. Again, the best solutions are not the results of a single algorithm. The bolded results in Table IV show the new best solutions that the RWPSO has produced. Table IV demonstrates how fast and consistently RWPSO responds to the problems of 100 customers. The typical variance is only a few percentage points, and the results seem to be very near to the optimal result. The results also point to consistency over repeated runs. The results showed that the RWPSO normally produces a respectable solution for the 100 customer problems because they originated from a variety of problem conditions.

The RWPSO performs better than [5] and [6] in exploring the atypical search space and provides competitive results that are distributed more uniformly across a variety of problem instances. Moreover, the RWPSO is effective at lowering the necessary number of vehicles. It can be shown that RWPSO is quite effective at handling situations with large vehicle capacities and time windows. This comparison demonstrates the RWPSO's effectiveness and competitiveness.

## VII. Conclusion

This paper applies a brand-new PSO version to VRPTW. Premature convergence is a problem with conventional PSO. With our new PSO methodology, the authors use the Roulette Wheel Method to prevent premature convergence. The proposed algorithm is known as RWPSO. Our algorithm's main benefit is that it prevents the best particle from falling into local minima by reducing the weight of the highest fitness solution and choosing the *gbest* using the Roulette Wheel Method instead. Computational experiments using the well-known Solomon datasets demonstrate that the RWPSO generates competitive results. Future work on this topic will explore whether this modification to PSO improves performance in other combinatorial optimization problems and whether alternative selection operators have a similar impact.

APPENDIX
Table II. 25 Customer Instances

| Instances | Total NV | Capacity | Solomon TD/Best Solution ([46]) | | RWPSO (best) | | RWPSO (Aggregated) | | | Average CPU Time | TD/Best Solution ([5]) | | TD/Best Solution ([6]) | | PSO | |
|---|---|---|---|---|---|---|---|---|---|---|---|---|---|---|---|---|
| | | | NV | TD | NV | TD | NV | TD (mean) | TD (std. dev) | | NV | TD | NV | TD | NV | TD |
| C101* | 25 | 200 | 3 | 191.3 | 3 | 191.81 | 3 | 192.82 | 23.13 | 37.96 | 3 | 191.81 | 3 | 191.81 | 3 | 191.8 |
| C102* | 25 | 200 | 3 | 190.3 | 3 | 190.72 | 3 | 190.73 | 3.45 | 81.66 | 3 | 190.74 | 3 | 190.73 | 3 | 190.7 |
| C103* | 25 | 200 | 3 | 190.3 | 3 | 190.72 | 3 | 190.73 | 2.11 | 39.39 | 3 | 190.74 | 3 | 190.73 | 3 | 190.7 |
| C104* | 25 | 200 | 3 | 186.9 | 3 | 186.95 | 3 | 186.96 | 9.03 | 36.52 | 3 | 187.45 | 3 | 188.52 | 3 | 192.1 |
| C105* | 25 | 200 | 3 | 191.3 | 3 | 191.81 | 3 | 192.9 | 19.47 | 38.89 | 3 | 191.81 | 3 | 190.73 | 3 | 191.8 |
| C106* | 25 | 200 | 3 | 191.3 | 3 | 191.81 | 3 | 192.9 | 22.75 | 39.93 | 3 | 191.81 | 3 | 190.73 | 3 | 191.8 |
| C107* | 25 | 200 | 3 | 191.3 | 3 | 191.81 | 3 | 192.9 | 21.16 | 37.07 | 3 | 191.81 | 3 | 190.73 | 3 | 191.8 |
| C108* | 25 | 200 | 3 | 191.3 | 3 | 191.81 | 3 | 192.9 | 18.78 | 36.99 | 3 | 191.81 | 3 | 190.73 | 3 | 191.8 |
| C109* | 25 | 200 | 3 | 191.3 | 3 | 191.81 | 3 | 192.9 | 14.87 | 38.34 | 3 | 191.81 | 3 | 190.73 | 3 | 191.8 |
| C201* | 25 | 700 | 2 | 214.7 | 2 | 215.54 | 2 | 216.55 | 8.64 | 80.54 | 2 | 215.54 | 2 | 215.54 | 2 | 215.5 |
| C202** | 25 | 700 | 2 | 214.7 | **1** | 222.32 | 1 | 223.4 | 0.00 | 81.04 | 1 | 223.31 | 1 | 215.54 | 2 | 215.5 |
| C203** | 25 | 700 | 2 | 214.7 | **1** | 222.32 | 1 | 223.4 | 0.54 | 79.48 | 1 | 223.31 | 1 | 215.54 | 2 | 215.5 |
| C204** | 25 | 700 | 2 | 213.1 | **1** | 222.18 | 1 | 223.19 | 1.22 | 78.85 | 1 | 221.28 | 1 | 213.93 | 2 | 213.9 |
| C205** | 25 | 700 | 2 | 214.7 | **1** | 216.15 | 1 | 217.16 | 7.34 | 80.21 | 1 | 297.45 | 1 | 215.54 | 2 | 215.5 |
| C206** | 25 | 700 | 2 | 214.7 | **1** | 216.29 | 1 | 217.3 | 7.34 | 79.75 | 1 | 285.39 | 1 | 215.54 | 2 | 215.5 |
| C207** | 25 | 700 | 2 | 214.5 | **1** | 214.78 | 1 | 215.79 | 5.77 | 78.63 | 1 | 274.28 | 1 | 215.33 | 2 | 215.3 |
| C208** | 25 | 700 | 2 | 214.5 | **1** | 219.14 | 1 | 220.15 | 5.38 | 78.12 | 1 | 229.84 | 1 | 215.54 | 2 | 215.4 |
| R101* | 25 | 200 | 8 | 617.1 | 8 | 618.31 | 8 | 619.32 | 61.81 | 95.06 | 8 | 618.33 | 8 | 618.32 | 8 | 618.3 |
| R102* | 25 | 200 | 7 | 547.1 | 7 | 548.11 | 7 | 549.12 | 29.04 | 100.78 | 7 | 548.11 | 7 | 548.1 | 7 | 548.1 |
| R103** | 25 | 200 | 5 | 454.6 | **4** | 455.19 | 4 | 456.2 | 34.49 | 91.4 | 4 | 473.39 | 4 | 455.69 | 5 | 459.2 |
| R104* | 25 | 200 | 4 | 416.9 | 4 | 417.19 | 4 | 418.2 | 20.62 | 87.51 | 4 | 418.39 | 4 | 417.96 | 4 | 418.1 |
| R105** | 25 | 200 | 6 | 530.5 | **5** | 536.22 | 5 | 537.23 | 16.00 | 95.00 | 5 | 556.72 | 6 | 531.53 | 6 | 531.5 |
| R106* | 25 | 200 | 5 | 465.4 | 5 | 467.18 | 5 | 468.19 | 181.09 | 90.04 | 5 | 466.48 | 5 | 468.95 | 3 | 466.5 |
| R107* | 25 | 200 | 4 | 424.3 | 4 | 425.26 | 4 | 426.27 | 162.78 | 88.69 | 4 | 425.27 | 4 | 425.26 | 4 | 426.8 |
| R108* | 25 | 200 | 4 | 397.3 | 4 | 401.49 | 4 | 402.5 | 173.18 | 88.41 | 4 | 405.39 | 4 | 398.29 | 4 | 398.3 |
| R109** | 25 | 200 | 5 | 441.3 | **4** | 444.22 | 4 | 445.23 | 7.41 | 94.90 | 4 | 460.52 | 5 | 442.62 | 5 | 442.6 |
| R110* | 25 | 200 | 4 | 441.1 | 4 | 429.33 | 4 | 430.34 | 18.20 | 94.49 | 4 | 429.7 | 4 | 429.69 | 4 | 449.9 |
| R111** | 25 | 200 | 5 | 428.8 | **4** | 429.33 | 4 | 430.34 | 9.26 | 94.04 | 4 | 429.7 | 4 | 429.69 | 5 | 432.9 |
| R112* | 25 | 200 | 4 | 393 | 4 | 394.1 | 4 | 395.2 | 15.89 | 87.8 | 4 | 394.1 | 4 | 394.1 | 4 | 394.1 |
| R201** | 25 | 1000 | 4 | 463.3 | **2** | 513.16 | 2 | 514.17 | 2.59 | 80.32 | 2 | 523.66 | 2 | 474.37 | 4 | 464.4 |
| R202** | 25 | 1000 | 4 | 410.5 | **2** | 415.23 | 2 | 416.24 | 3.28 | 80.33 | 2 | 455.53 | 2 | 427.11 | 4 | 411.5 |
| R203** | 25 | 1000 | 3 | 391.4 | **2** | 402.19 | 2 | 403.2 | 3.03 | 81.35 | 2 | 408.89 | 2 | 406.69 | 3 | 392.3 |
| R204** | 25 | 1000 | 2 | 355 | **1** | 369.94 | 1 | 370.95 | 5.03 | 78.82 | 1 | 389.91 | 1 | 356.61 | 2 | 358.6 |
| R205** | 25 | 1000 | 3 | 393 | **1** | 406.13 | 1 | 407.14 | 4.02 | 72.72 | 1 | 501.83 | 1 | 405.97 | 3 | 395.8 |
| R206** | 25 | 1000 | 3 | 374.4 | **1** | 400.11 | 1 | 401.12 | 4.46 | 72.23 | 1 | 413.21 | 1 | 380.75 | 3 | 378.8 |
| R207** | 25 | 1000 | 3 | 361.6 | **1** | 370.18 | 1 | 371.19 | 2.65 | 72.14 | 1 | 402.28 | 1 | 360.79 | 3 | 362.6 |
| R208* | 25 | 1000 | 1 | 328.2 | 1 | 329.32 | 1 | 330.33 | 3.26 | 71.72 | 1 | 329.33 | 1 | 332.01 | 1 | 329.3 |
| R209** | 25 | 1000 | 2 | 370.7 | **1** | 418.21 | 1 | 419.22 | 2.67 | 73.05 | 1 | 438.24 | 1 | 371.56 | 2 | 371.6 |
| R210** | 25 | 1000 | 3 | 404.6 | **1** | 413.98 | 1 | 414.99 | 4.01 | 73.42 | 1 | 513.98 | 1 | 413.44 | 3 | 405.5 |
| R211** | 25 | 1000 | 2 | 350.9 | **1** | 351.19 | 1 | 352.2 | 2.46 | 71.79 | 1 | 361.69 | 1 | 351.91 | 2 | 353.8 |
| RC101* | 25 | 200 | 4 | 461.1 | 4 | 462.16 | 4 | 463.17 | 5.45 | 93.26 | 4 | 462.16 | 4 | 462.15 | 4 | 462.2 |
| RC102* | 25 | 200 | 3 | 351.8 | 3 | 352.74 | 3 | 353.75 | 416.51 | 92.53 | 3 | 352.74 | 3 | 352.74 | 3 | 352.7 |
| RC103* | 25 | 200 | 3 | 332.8 | 3 | 333.92 | 3 | 334.93 | 11.30 | 92.14 | 3 | 333.92 | 3 | 333.91 | 3 | 333.9 |
| RC104* | 25 | 200 | 3 | 306.6 | 3 | 307.14 | 3 | 308.15 | 663.46 | 90.68 | 3 | 307.14 | 3 | 307.13 | 3 | 307.1 |
| RC105* | 25 | 200 | 4 | 411.3 | 4 | 412.34 | 4 | 413.35 | 234.51 | 93.15 | 4 | 412.38 | 4 | 412.37 | 4 | 412.4 |
| RC106* | 25 | 200 | 3 | 345.5 | 3 | 346.51 | 3 | 347.52 | 0 | 86.32 | 3 | 346.51 | 3 | 346.5 | 3 | 347.1 |
| RC107* | 25 | 200 | 3 | 298.3 | 3 | 298.95 | 3 | 299.96 | 12.34 | 90.31 | 3 | 298.95 | 3 | 298.94 | 3 | 298.9 |
| RC108* | 25 | 200 | 3 | 294.5 | 3 | 294.99 | 3 | 295.1 | 36.02 | 90.08 | 3 | 294.99 | 3 | 294.99 | 3 | 295.1 |
| RC201** | 25 | 1000 | 3 | 360.2 | **2** | 432.33 | 2 | 433.34 | 1.73 | 75.94 | 2 | 432.3 | 2 | 361.24 | 3 | 361.2 |
| RC202** | 25 | 1000 | 3 | 338 | **2** | 356.14 | 2 | 357.15 | 9.92 | 75.13 | 2 | 376.12 | 1 | 338.82 | 3 | 338.8 |
| RC203** | 25 | 1000 | 3 | 326.9 | **1** | 332.45 | 1 | 333.46 | 9.28 | 75.61 | 1 | 432.55 | 1 | 328.44 | 3 | 327.7 |
| RC204** | 25 | 1000 | 3 | 299.7 | **1** | 347.23 | 1 | 348.24 | 6.75 | 74.64 | 1 | 327.33 | 1 | 300.98 | 3 | 300.2 |
| RC205** | 25 | 1000 | 3 | 338 | **2** | 346.65 | 2 | 347.66 | 25.43 | 86.44 | 2 | 386.15 | 2 | 338.92 | 3 | 338.9 |
| RC206** | 25 | 1000 | 3 | 324 | **1** | 422.62 | 1 | 423.63 | 1.66 | 76.65 | 1 | 482.02 | 1 | 325.1 | 3 | 325.1 |
| RC207** | 25 | 1000 | 3 | 298.3 | **1** | 378.17 | 1 | 379.18 | 24.37 | 76.15 | 1 | 478.97 | 1 | 229.49 | 3 | 228.9 |
| RC208** | 25 | 1000 | 2 | 269.1 | **1** | 409.65 | 1 | 410.66 | 2.43 | 74.77 | 1 | 309.85 | 1 | 269.56 | 2 | 269.6 |

(*) – 28; (**) – 28

Table III. 50 Customer Instances

| Instances | Total NV | Capacity | Solomon TD/Best Solution ([46]) | | RWPSO (best) | | RWPSO (Aggregated) | | | Average CPU Time | TD/Best Solution ([5]) | | TD/Best Solution ([6]) | | PSO | |
|---|---|---|---|---|---|---|---|---|---|---|---|---|---|---|---|---|
| | | | NV | TD | NV | TD | NV (mean) | TD | TD (std. dev) | | NV | TD | NV | TD | NV | TD |
| C101* | 25 | 200 | 5 | 362.4 | 5 | 363.15 | 5 | 364.16 | 252.6 | 368.42 | 5 | 363.25 | 5 | 363.2 | 5 | 377.5 |
| C102* | 25 | 200 | 5 | 361.4 | 5 | 362.17 | 5 | 363.18 | 153.4 | 361.41 | 5 | 362.17 | 5 | 36.17 | 5 | 408.9 |
| C103* | 25 | 200 | 5 | 361.4 | 5 | 363.27 | 5 | 364.28 | 196.9 | 379.01 | 5 | 362.17 | 5 | 369.2 | 5 | 430.9 |
| C104* | 25 | 200 | 5 | 358 | 5 | 359.85 | 5 | 360.86 | 311.5 | 377.87 | 5 | 358.88 | 5 | 363.9 | 5 | 389.1 |
| C105* | 25 | 200 | 5 | 362.4 | 5 | 363.55 | 5 | 364.56 | 287.8 | 363.41 | 5 | 363.25 | 5 | 363.2 | 5 | 383.9 |
| C106* | 25 | 200 | 5 | 362.4 | 5 | 363.55 | 5 | 364.56 | 99.61 | 362.98 | 5 | 363.25 | 5 | 363.2 | 5 | 370.3 |
| C107* | 25 | 200 | 5 | 362.4 | 5 | 363.55 | 5 | 364.56 | 288.0 | 371.29 | 5 | 363.25 | 5 | 363.2 | 5 | 406.5 |
| C108* | 25 | 200 | 5 | 362.4 | 5 | 363.55 | 5 | 364.56 | 234.6 | 359.38 | 5 | 363.25 | 5 | 363.2 | 5 | 403.4 |
| C109* | 25 | 200 | 5 | 362.4 | 5 | 363.55 | 5 | 364.56 | 176.2 | 368.12 | 5 | 363.25 | 5 | 363.2 | 5 | 420.1 |
| C201** | 25 | 700 | 3 | 360.2 | **2** | 424.46 | 2 | 425.47 | 35.89 | 293.81 | 2 | 444.96 | 2 | 361.7 | 3 | 361.8 |
| C202** | 25 | 700 | 3 | 360.2 | **2** | 433.21 | 2 | 434.22 | 111.4 | 302.19 | 2 | 403.81 | 2 | 361.7 | 3 | 361.8 |
| C203** | 25 | 700 | 3 | 359.8 | **2** | 412.32 | 2 | 413.33 | 96.03 | 305.01 | 2 | 402.52 | 2 | 370.9 | 2 | 361.4 |
| C204* | 25 | 700 | 2 | 350.1 | 2 | 376.47 | 2 | 377.48 | 154.3 | 292.76 | 2 | 356.77 | 2 | 364.1 | 2 | 351.7 |
| C205** | 25 | 700 | 3 | 359.8 | **2** | 419.15 | 2 | 420.16 | 198.2 | 287.3 | 2 | 429.12 | 2 | 361.4 | 3 | 361.4 |
| C206** | 25 | 700 | 3 | 359.8 | **2** | 414.1 | 2 | 415.2 | 61.98 | 294.88 | 2 | 412.5 | 2 | 361.4 | 3 | 361.4 |
| C207** | 25 | 700 | 3 | 359.6 | **2** | 456.14 | 2 | 457.15 | 133.0 | 289.06 | 2 | 426.13 | 2 | 361.2 | 3 | 361.2 |
| C208* | 25 | 700 | 2 | 350.5 | 2 | 352.59 | 2 | 353.6 | 42.14 | 292.07 | 2 | 352.29 | 2 | 352.1 | 2 | 352.1 |
| R101** | 25 | 200 | 12 | 1044 | **11** | 1090.42 | 11 | 1091.43 | 362.35 | 450.16 | 11 | 1100.72 | 11 | 1052 | 12 | 1054.3 |
| R102** | 25 | 200 | 11 | 909 | **10** | 913.21 | 10 | 914.22 | 251.1 | 452.87 | 10 | 923.71 | 10 | 914.4 | 11 | 918.8 |
| R103** | 25 | 200 | 9 | 772.9 | **8** | 794.67 | 8 | 795.68 | 298.4 | 431.14 | 8 | 790.17 | 8 | 782.7 | 9 | 780.8 |
| R104* | 25 | 200 | 6 | 625.4 | 6 | 635.48 | 6 | 636.49 | 301.7 | 442.98 | 6 | 631.58 | 6 | 649.7 | 6 | 634.9 |
| R105* | 25 | 200 | 9 | 899.3 | **8** | 933.45 | 8 | 934.46 | 376.3 | 437.05 | 8 | 983.49 | 9 | 918.5 | 9 | 945 |
| R106* | 25 | 200 | 5 | 793 | 5 | 835.91 | 7 | 836.92 | 221.5 | 422.21 | 7 | 865.93 | 8 | 809.5 | 9 | 810.3 |
| R107** | 25 | 200 | 7 | 711.1 | **6** | 727.13 | 6 | 728.14 | 403.9 | 430.03 | 6 | 737.1 | 7 | 747.9 | 7 | 732.7 |
| R108* | 25 | 200 | 6 | 617.7 | 6 | 619.19 | 6 | 620.2 | 329.0 | 444.78 | 6 | 624.29 | 6 | 635.4 | 6 | 620.7 |
| R109** | 25 | 200 | 8 | 786.8 | **7** | 805.92 | 7 | 806.93 | 198.4 | 453.47 | 7 | 801.97 | 8 | 806.5 | 8 | 817.2 |
| R110* | 25 | 200 | 7 | 697 | 7 | 723.43 | 7 | 724.44 | 341.3 | 449.26 | 7 | 720.4 | 7 | 725.3 | 7 | 731.8 |
| R111** | 25 | 200 | 7 | 707.2 | **6** | 716.15 | 6 | 717.16 | 276.8 | 456.14 | 7 | 756.35 | 7 | 725.9 | 7 | 728 |
| R112* | 25 | 200 | 6 | 630.2 | 6 | 648.19 | 6 | 649.2 | 257.5 | 443.35 | 6 | 638.49 | 6 | 660.3 | 6 | 642.6 |
| R201** | 25 | 1000 | 6 | 791.9 | **2** | 913.25 | 2 | 914.26 | 28.1 | 317.93 | 2 | 953.29 | 3 | 811.1 | 6 | 815.7 |
| R202** | 25 | 1000 | 5 | 698.5 | **2** | 725.13 | 2 | 726.14 | 21.65 | 318.76 | 2 | 815.23 | 2 | 716.2 | 5 | 725.9 |
| R203** | 25 | 1000 | 5 | 605.3 | **2** | 658.37 | 2 | 659.38 | 97.34 | 314.91 | 2 | 668.36 | 2 | 623.3 | 5 | 616.6 |
| R204* | 25 | 1000 | 2 | 506.4 | 2 | 511.27 | 2 | 512.28 | 58.68 | 309.58 | 2 | 518.57 | 2 | 512.9 | 2 | 507.1 |
| R205** | 25 | 1000 | 4 | 690.1 | **2** | 716.28 | 2 | 717.29 | 104.5 | 311.71 | 2 | 756.38 | 2 | 703.2 | 4 | 709 |
| R206** | 25 | 1000 | 4 | 632.4 | **2** | 663.15 | 2 | 664.16 | 11.02 | 321.34 | 2 | 661.55 | 2 | 648.2 | 4 | 638.6 |
| R207* | 25 | 1000 | 2 | 593.9 | 2 | 595.15 | 2 | 596.16 | 22.96 | 312.06 | 2 | 593.95 | 2 | 596.3 | 2 | 598.8 |
| R208* | 25 | 1000 | 2 | 508.4 | 2 | 509.61 | 2 | 510.62 | 70.11 | 317.57 | 2 | 508.41 | 2 | 501 | 2 | 505 |
| R209** | 25 | 1000 | 4 | 600.6 | **2** | 628.48 | 2 | 629.49 | 45.34 | 322.18 | 2 | 658.28 | 2 | 622.9 | 4 | 624.6 |
| R210** | 25 | 1000 | 4 | 645.6 | **2** | 660.69 | 2 | 661.7 | 20.39 | 316.11 | 2 | 670.99 | 2 | 651.4 | 4 | 657.8 |
| R211** | 25 | 1000 | 3 | 535.5 | **2** | 552.14 | 2 | 553.15 | 12.83 | 318.6 | 2 | 562.74 | 2 | 554.8 | 3 | 539.2 |
| RC101* | 25 | 200 | 8 | 944 | 8 | 946.5 | 8 | 947.6 | 775.3 | 398.32 | 8 | 945.58 | 8 | 946.4 | 8 | 947.9 |
| RC102* | 25 | 200 | 7 | 822.5 | 7 | 843.44 | 7 | 844.43 | 666.8 | 385.83 | 7 | 823.97 | 7 | 846.8 | 7 | 830.3 |
| RC103* | 25 | 200 | 6 | 710.9 | 6 | 742.36 | 6 | 743.37 | 702.2 | 382.19 | 6 | 712.91 | 7 | 761.1 | 6 | 713.1 |
| RC104* | 25 | 200 | 5 | 545.8 | 5 | 547.11 | 5 | 548.12 | 744.0 | 379.06 | 5 | 546.91 | 5 | 554.0 | 5 | 546.5 |
| RC105* | 25 | 200 | 8 | 855.3 | 8 | 866.17 | 8 | 867.18 | 812.4 | 402.51 | 8 | 856.97 | 8 | 879.7 | 8 | 860.9 |
| RC106* | 25 | 200 | 6 | 723.2 | 6 | 725.64 | 6 | 726.65 | 695.7 | 391.97 | 6 | 724.65 | 6 | 775.4 | 6 | 761.6 |
| RC107* | 25 | 200 | 6 | 642.7 | 6 | 643.74 | 6 | 644.75 | 798.1 | 385.12 | 6 | 645.7 | 6 | 647.8 | 6 | 645.5 |
| RC108* | 25 | 200 | 6 | 598.1 | 6 | 599.13 | 6 | 5100.1 | 652.6 | 384.88 | 6 | 599.17 | 6 | 604.4 | 6 | 599.2 |
| RC201** | 25 | 1000 | 5 | 684.8 | **3** | 808.26 | 3 | 809.27 | 71.23 | 319.31 | 3 | 838.76 | 3 | 686.3 | 5 | 686.3 |
| RC202** | 25 | 1000 | 5 | 613.6 | **2** | 827.46 | 2 | 828.47 | 85.98 | 329.98 | 2 | 867.26 | 2 | 617.4 | 5 | 615.0 |
| RC203** | 25 | 1000 | 4 | 555.3 | **2** | 614.49 | 2 | 615.51 | 112.7 | 327.37 | 2 | 674.44 | 2 | 571.1 | 4 | 556.5 |
| RC204* | 25 | 1000 | 3 | 444.2 | **2** | 449.28 | 2 | 450.29 | 52.56 | 316.29 | 2 | 479.22 | 2 | 448.6 | 3 | 445.0 |
| RC205* | 25 | 1000 | 5 | 630.2 | **3** | 735.01 | 3 | 736.02 | 39.98 | 322.94 | 3 | 765.02 | 3 | 633.5 | 5 | 632.0 |
| RC206* | 25 | 1000 | 5 | 610 | **2** | 705.14 | 2 | 706.15 | 102.6 | 318.65 | 2 | 755.13 | 2 | 612.5 | 5 | 611.7 |
| RC207* | 25 | 1000 | 4 | 558.6 | **2** | 685.8 | 2 | 686.83 | 79.02 | 316.87 | 2 | 655.81 | 2 | 568.4 | 4 | 559.9 |
| RC208* | 25 | 1000 | 2 | 498.7 | 2 | 498.9 | 2 | 499.9 | 61.3 | 314.3 | 2 | 498.9 | 2 | 491.25 | 2 | 495.2 |

(*) – 29; (**) – 27

Table IV. 100 Customer Instances

| Instances | Total NV | Capacity | Solomon TD/Best Solution ([46]) | | RWPSO (best) | | RWPSO (Aggregated) | | | Average CPU Time | TD/Best Solution ([5]) | | TD/Best Solution ([6]) | | PSO | |
|---|---|---|---|---|---|---|---|---|---|---|---|---|---|---|---|---|
| | | | NV | TD | NV | TD | NV (mean) | TD | TD (std. dev) | | NV | TD | NV | TD | NV | TD |
| C101** | 25 | 200 | 10 | 828.94 | **9** | 828.93 | 10 | 829.95 | 152.7 | 1598.6 | 10 | 828.93 | 10 | 828.93 | 10 | 828.94 |
| C102* | 25 | 200 | 10 | 828.94 | 10 | 829.61 | 10 | 830.63 | 328.0 | 1732.6 | 10 | 829.71 | 10 | 828.93 | 10 | 828.94 |
| C103* | 25 | 200 | 10 | 828.06 | 10 | 831.27 | 10 | 832.28 | 466.2 | 1780.8 | 10 | 851.37 | 10 | 828.06 | 10 | 828.06 |
| C104* | 25 | 200 | 10 | 824.78 | 10 | 838.32 | 10 | 839.34 | 397.5 | 1759.4 | 10 | 868.52 | 10 | 824.77 | 10 | 824.78 |
| C105** | 25 | 200 | 10 | 828.94 | **9** | 828.93 | 10 | 829.94 | 233.4 | 1564.1 | 10 | 828.93 | 10 | 828.93 | 10 | 828.94 |
| C106* | 25 | 200 | 10 | 828.94 | 10 | 828.93 | 10 | 829.94 | 192.8 | 1582.2 | 10 | 828.93 | 10 | 828.93 | 10 | 828.94 |
| C107** | 25 | 200 | 10 | 828.94 | **9** | 828.93 | 10 | 829.94 | 211.6 | 1638.9 | 10 | 828.93 | 10 | 828.93 | 10 | 828.94 |
| C108** | 25 | 200 | 10 | 828.94 | **9** | 828.93 | 10 | 829.94 | 174.2 | 1599.3 | 10 | 828.93 | 10 | 828.93 | 10 | 828.94 |
| C109* | 25 | 200 | 10 | 828.94 | 10 | 828.93 | 10 | 829.94 | 188.3 | 1553.2 | 10 | 828.93 | 10 | 828.93 | 10 | 828.94 |
| C201* | 25 | 700 | 3 | 591.56 | 3 | 591.55 | 3 | 592.56 | 117.8 | 1281.0 | 3 | 591.55 | 3 | 591.55 | 3 | 591.56 |
| C202* | 25 | 700 | 3 | 591.56 | 3 | 591.55 | 3 | 592.56 | 165.3 | 1272.7 | 3 | 591.55 | 3 | 591.55 | 3 | 591.56 |
| C203** | 25 | 700 | 3 | 591.17 | **2** | 591.17 | 3 | 592.18 | 149.9 | 1409.6 | 3 | 591.17 | 3 | 591.17 | 3 | 591.17 |
| C204** | 25 | 700 | 3 | 590.6 | **2** | 617.13 | 3 | 618.14 | 141.1 | 1301.8 | 3 | 615.43 | 3 | 590.99 | 3 | 590.6 |
| C205** | 25 | 700 | 3 | 588.88 | **2** | 588.87 | 3 | 589.88 | 109.7 | 1239.2 | 3 | 588.87 | 3 | 588.87 | 3 | 588.88 |
| C206* | 25 | 700 | 3 | 588.49 | 3 | 588.57 | 3 | 589.58 | 98.83 | 1215.9 | 3 | 588.87 | 3 | 588.49 | 3 | 588.49 |
| C207* | 25 | 700 | 3 | 588.29 | 3 | 593.32 | 3 | 594.33 | 123.2 | 1327.4 | 3 | 591.35 | 3 | 588.28 | 3 | 588.29 |
| C208* | 25 | 700 | 3 | 588.32 | 3 | 588.29 | 3 | 589.3 | 111.6 | 1254.0 | 3 | 588.49 | 3 | 588.32 | 3 | 588.32 |
| R101** | 25 | 200 | 19 | 1645.7 | **18** | 1651 | 19 | 1652 | 419.3 | 2189.7 | 19 | 1652 | 19 | 1650.7 | 20 | 1642.8 |
| R102** | 25 | 200 | 17 | 1486.1 | **16** | 1520.3 | 17 | 1521.4 | 501.7 | 2292.2 | 17 | 1500.8 | 17 | 1486.8 | 18 | 1472.6 |
| R103* | 25 | 200 | 13 | 1292.6 | 13 | 1293.1 | 14 | 1294.1 | 432.9 | 1976.6 | 14 | 1242.6 | 13 | 1241.3 | 14 | 1213.6 |
| R104* | 25 | 200 | 9 | 1007.2 | 9 | 1022.1 | 10 | 1023.1 | 345.8 | 1733.1 | 10 | 1042.2 | 10 | 1007.3 | 9 | 1007.2 |
| R105* | 25 | 200 | 14 | 1377.1 | 14 | 1381.1 | 14 | 1382.1 | 311.6 | 1831.8 | 14 | 1385.0 | 14 | 1377.1 | 15 | 1360.7 |
| R106* | 25 | 200 | 12 | 1251.9 | 12 | 1264.2 | 12 | 1265.2 | 490.2 | 1855.3 | 12 | 1294.8 | 12 | 1252.0 | 13 | 1241.5 |
| R107* | 25 | 200 | 10 | 1104.6 | 10 | 1113.2 | 11 | 1114.2 | 432.9 | 1794.5 | 11 | 1123.9 | 10 | 1104.6 | 11 | 1076.1 |
| R108* | 25 | 200 | 9 | 960.88 | 9 | 1001.1 | 10 | 1002.1 | 408.7 | 1708.8 | 10 | 1011.6 | 10 | 960.87 | 9 | 963.99 |
| R109* | 25 | 200 | 11 | 1194.7 | 11 | 1241.1 | 12 | 1242.1 | 419.0 | 1697.7 | 12 | 1211.6 | 12 | 1211.6 | 13 | 1151.8 |
| R110* | 25 | 200 | 10 | 1118.5 | 10 | 1120.1 | 11 | 1121.1 | 422.4 | 1601.9 | 11 | 1190.3 | 11 | 1190.8 | 11 | 1080.3 |
| R111* | 25 | 200 | 10 | 1096.7 | **9** | 1132.1 | 11 | 1133.1 | 495.5 | 1783.4 | 11 | 1102.9 | 11 | 1102.7 | 12 | 1053.5 |
| R112* | 25 | 200 | 9 | 982.14 | 9 | 1019.2 | 10 | 1020.2 | 518.8 | 1586.3 | 10 | 1029.1 | 10 | 982.14 | 10 | 953.63 |
| R201* | 25 | 1000 | 4 | 1252.3 | 4 | 1254.1 | 4 | 1255.1 | 397.5 | 1274.4 | 4 | 1274.7 | 4 | 1182.2 | 9 | 1179.7 |
| R202* | 25 | 1000 | 3 | 1191.7 | 3 | 1217.1 | 3 | 1218.1 | 510.2 | 1359.9 | 3 | 1247.0 | 4 | 1191.7 | 7 | 1049.7 |
| R203* | 25 | 1000 | 3 | 939.54 | 3 | 1012.7 | 3 | 1013.7 | 452.9 | 1368.3 | 3 | 1052.7 | 3 | 939.5 | 7 | 932.76 |
| R204** | 25 | 1000 | 2 | 825.52 | **1** | 841.14 | 3 | 842.15 | 401.2 | 1225.8 | 3 | 844.16 | 3 | 825.51 | 4 | 772.33 |
| R205* | 25 | 1000 | 3 | 994.42 | 3 | 1031.4 | 3 | 1032.4 | 385.3 | 1291.7 | 3 | 1061.4 | 3 | 994.42 | 6 | 970.89 |
| R206* | 25 | 1000 | 3 | 906.14 | 3 | 1026.3 | 3 | 1027.3 | 448.3 | 1264.9 | 3 | 1016.3 | 3 | 906.71 | 3 | 906.14 |
| R207* | 25 | 1000 | 2 | 893.33 | 2 | 941.72 | 3 | 942.73 | 462.5 | 1277.1 | 3 | 946.77 | 3 | 890.6 | 3 | 814.78 |
| R208* | 25 | 1000 | 2 | 726.75 | 2 | 834.12 | 2 | 835.13 | 531.7 | 1210.6 | 2 | 834.72 | 2 | 726.82 | 4 | 725.42 |
| R209* | 25 | 1000 | 3 | 909.16 | 3 | 1013.1 | 3 | 1014.1 | 423.9 | 1223.8 | 3 | 1003.1 | 3 | 909.16 | 6 | 879.53 |
| R210* | 25 | 1000 | 3 | 939.34 | 3 | 1010.5 | 3 | 1011.5 | 586.8 | 1276.1 | 3 | 1040.5 | 3 | 939.37 | 3 | 954.12 |
| R211* | 25 | 1000 | 2 | 892.71 | 2 | 895.34 | 3 | 896.35 | 443.3 | 1209.1 | 3 | 861.32 | 3 | 885.71 | 2 | 885.71 |
| RC101* | 25 | 200 | 14 | 1696.9 | 14 | 1691.2 | 15 | 1692.2 | 345.9 | 1878.4 | 15 | 1641.2 | 15 | 1668.9 | 15 | 1623.5 |
| RC102** | 25 | 200 | 12 | 1554.7 | **11** | 1560.4 | 13 | 1562.4 | 318.2 | 1814.8 | 13 | 1510.9 | 13 | 1510.7 | 14 | 1482.9 |
| RC103** | 25 | 200 | 11 | 1261.6 | **10** | 1274.7 | 11 | 1275.72 | 238.21 | 1696.17 | 11 | 1294.73 | 11 | 1261.67 | 11 | 1262.02 |
| RC104* | 25 | 200 | 10 | 1135.48 | **9** | 1193.51 | 10 | 1194.52 | 287.93 | 1637.74 | 10 | 1190.54 | 10 | 1135.47 | 10 | 1135.48 |
| RC105* | 25 | 200 | 13 | 1629.4 | 13 | 1643.7 | 14 | 1644.7 | 303.6 | 1759.2 | 14 | 1603.7 | 13 | 1594.1 | 16 | 1518.6 |
| RC106* | 25 | 200 | 11 | 1424.7 | 11 | 1428.9 | 12 | 1429.9 | 299.0 | 1666.0 | 12 | 1410.9 | 12 | 1410.7 | 12 | 1384.9 |
| RC107* | 25 | 200 | 11 | 1230.4 | 11 | 1249.1 | 11 | 1250.2 | 226.8 | 1679.7 | 11 | 1249.7 | 11 | 1230.4 | 12 | 1212.8 |
| RC108** | 25 | 200 | 10 | 1139.8 | **9** | 1151.8 | 11 | 1152.8 | 247.9 | 1623.9 | 11 | 1181.8 | 11 | 1139.8 | 11 | 1117.5 |
| RC201** | 25 | 1000 | 4 | 1406.9 | **3** | 1443.5 | 4 | 1444.5 | 394.1 | 1274.9 | 4 | 1423.5 | 4 | 1314.3 | 4 | 1406.9 |
| RC202* | 25 | 1000 | 3 | 1367.0 | 3 | 1393.2 | 4 | 1394.3 | 358.7 | 1198.3 | 4 | 1193.5 | 4 | 1203.6 | 8 | 1113.5 |
| RC203* | 25 | 1000 | 3 | 1049.6 | 3 | 1121.4 | 3 | 1122.4 | 308.9 | 1262.8 | 3 | 1123.4 | 3 | 966.73 | 5 | 945.69 |
| RC204* | 25 | 1000 | 3 | 798.41 | 3 | 891.12 | 3 | 892.13 | 275.0 | 1207.4 | 3 | 894.11 | 3 | 798.46 | 3 | 798.41 |
| RC205* | 25 | 1000 | 4 | 1297.1 | 4 | 1320.4 | 4 | 1321.4 | 318.2 | 1288.5 | 4 | 1321.4 | 4 | 1241.3 | 8 | 1168.2 |
| RC206* | 25 | 1000 | 3 | 1146.3 | 3 | 1207.1 | 3 | 1208.1 | 301.6 | 1217.6 | 3 | 1307.9 | 3 | 1142.4 | 8 | 1084.3 |
| RC207* | 25 | 1000 | 3 | 1061.1 | 3 | 1230.1 | 3 | 1231.1 | 299.2 | 1302.9 | 3 | 1130.3 | 3 | 1061.1 | 7 | 976.4 |
| RC208* | 25 | 1000 | 3 | 828.14 | 3 | 928.22 | 3 | 929.23 | 263.5 | 1277.1 | 3 | 958.23 | 3 | 828.14 | 5 | 816.1 |

(*) – 41; (**) – 15;